\documentclass[conference]{IEEEtran}
\usepackage{authblk}
\usepackage{epsfig}
\usepackage{multirow}
\usepackage{wrapfig}
\usepackage{hyperref}
\usepackage{listings}
\usepackage{amsbsy}
\usepackage{color}
\usepackage{amssymb}
\usepackage{amsmath} % need this for equation array in Sol 2 sec
\usepackage{tabularx} % need this for equation array in Sol 2 sec
\usepackage{booktabs}
\usepackage{multirow}
\usepackage[utf8]{inputenc}
\usepackage[english]{babel}

\newcommand{\comment}[1]

\ifCLASSINFOpdf
\else
\fi
\title{Using Non-invertible Data Transformations to Build Adversarial-Robust Neural Networks}
\author[1,2]{Qinglong Wang}
\author[1,3]{Wenbo Guo}
\author[1]{Alexander G. Ororbia II}
\author[1]{Xinyu Xing}
\author[1]{Lin Lin}
\author[1]{C. Lee Giles}
\author[2]{\\Xue Liu}
\author[1]{Peng Liu}
\author[4]{Gang Xiong}

\affil[1]{Pennsylvania State University} 
\affil[2]{McGill University}
\affil[3]{Shanghai Jiao Tong University}
\affil[4]{Chinese Academy of Sciences}

\begin{document}

% author names and affiliations
% use a multiple column layout for up to three different
% affiliations
%\author{\IEEEauthorblockN{Qinglong Wang}
%\IEEEauthorblockA{McGill University\\
%qinglong.wang@mail.mcgill.ca}
%\and
%\IEEEauthorblockN{Wenbo Guo}
%\IEEEauthorblockA{Shanghai Jiao Tong University\\
%henrygwb@gmail.com}}

% conference papers do not typically use \thanks and this command
% is locked out in conference mode. If really needed, such as for
% the acknowledgment of grants, issue a \IEEEoverridecommandlockouts
% after \documentclass

% for over three affiliations, or if they all won't fit within the width
% of the page, use this alternative format:
%
%\author{\IEEEauthorblockN{Michael Shell\IEEEauthorrefmark{1},
%Homer Simpson\IEEEauthorrefmark{2},
%James Kirk\IEEEauthorrefmark{3},
%Montgomery Scott\IEEEauthorrefmark{3} and
%Eldon Tyrell\IEEEauthorrefmark{4}}
%\IEEEauthorblockA{\IEEEauthorrefmark{1}School of Electrical and Computer Engineering\\
%Georgia Institute of Technology,
%Atlanta, Georgia 30332--0250\\ Email: see http://www.michaelshell.org/contact.html}
%\IEEEauthorblockA{\IEEEauthorrefmark{2}Twentieth Century Fox, Springfield, USA\\
%Email: homer@thesimpsons.com}
%\IEEEauthorblockA{\IEEEauthorrefmark{3}Starfleet Academy, San Francisco, California 96678-2391\\
%Telephone: (800) 555--1212, Fax: (888) 555--1212}
%\IEEEauthorblockA{\IEEEauthorrefmark{4}Tyrell Inc., 123 Replicant Street, Los Angeles, California 90210--4321}}

% use for special paper notices
%\IEEEspecialpapernotice{(Invited Paper)}

\IEEEoverridecommandlockouts
\makeatletter\def\@IEEEpubidpullup{9\baselineskip}\makeatother
\IEEEpubid{\parbox{\columnwidth}{Permission to freely reproduce all or part
    of this paper for noncommercial purposes is granted provided that
    copies bear this notice and the full citation on the first
    page. Reproduction for commercial purposes is strictly prohibited
    without the prior written consent of the Internet Society, the
    first-named author (for reproduction of an entire paper only), and
    the author's employer if the paper was prepared within the scope
    of employment.
}
\hspace{\columnsep}\makebox[\columnwidth]{}}

% make the title area
\maketitle
\begin{abstract}
Deep neural networks have proven to be quite effective in a wide variety of machine learning tasks, ranging from improved speech recognition systems to advancing the development of autonomous vehicles. However, despite their superior performance in many applications, these models have been recently shown to be susceptible to a particular type of attack possible through the generation of particular synthetic examples referred to as adversarial samples. These samples are constructed by manipulating real examples from the training data distribution in order to ``\emph{fool}'' the original neural model, resulting in misclassification (with high confidence) of previously correctly classified samples. Addressing this weakness is of utmost importance if deep neural architectures are to be applied to critical applications, such as those in the domain of cybersecurity. In this paper, we present an analysis of this fundamental flaw lurking in all neural architectures to uncover limitations of previously proposed defense mechanisms. More importantly, we present a unifying framework for protecting deep neural models using a non-invertible data transformation--developing two  adversary-resilient architectures utilizing both linear and nonlinear dimensionality reduction. Empirical results indicate that our framework provides better robustness compared to state-of-art solutions while having negligible degradation in accuracy.    
\end{abstract}
\IEEEpeerreviewmaketitle
\section{Introduction}
\label{sec:intro}
 
Aside from its highly publicized victories in Go~\cite{silver2016mastering}, there have been numerous successful applications of deep neural network (DNN) learning in image and speech recognition. Recent interest has been to integrate it into critical fields like medical imaging~\cite{bar2015deep, xu2014deep} and self-driving cars~\cite{hadsell2009learning, farabet2012scene}. In cybersecurity, security companies have demonstrated that deep learning could offer a far better way to detect all types of malware~\cite{pascanu2015malware, dahl2013large, yuan2014droid} and recognize such functions in binary code~\cite{shin2015recognizing}.

\begin{figure*}[t]
\begin{center}
  \includegraphics[scale=0.60]{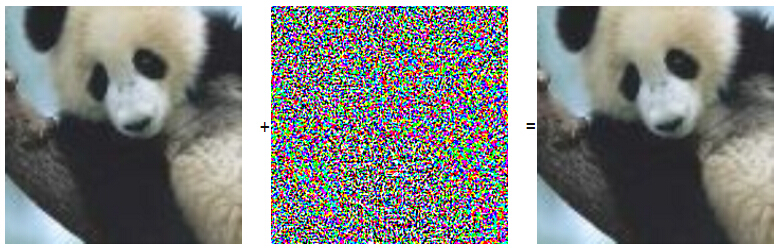}\\
  \caption{Demonstration of an adversarial example generated from a panda picture~\cite{Goodfellow14}. The left picture represents the original image classified as a panda with $60\%$ confidence. Right picture is an adversarial sample obtained by adding a tiny perturbation to the original picture. Despite being visually similar to the original image, the adversarial sample is classified as a gibbon with $99\%$ confidence.}
  \label{fig:panda}
\end{center}
\end{figure*}

However, recent work\cite{42503, nguyen2015deep} uncovered a potential flaw in DNN-powered systems that could be readily exploited by an attack.  They show that an attacker could use the same learning algorithm, back-propagation, and a surrogate data set to construct an auxiliary model that can accurately approximate the target DNN model. When compared to the original model that usually returns categorical classification results (e.g., benign or malicious binary code), such an auxiliary model could provide the attacker with useful details about a DNN's weaknesses (e.g., a continuous classification score for a malware sample). As such, the attacker can use the auxiliary model to perform invertible computation, or examine class/feature importance to identify the features that have significant impact on the target's classification ability. With this knowledge of feature importance, the attacker can minimize effort in crafting an {\em adversarial sample} -- a synthetic example generated by modifying a real example slightly in order to make the original DNN model believe it belongs to the wrong class with high confidence. For example, Fig.~\ref{fig:panda} illustrates the case where an adversarial sample constructed by modifying a picture of a panda misleads the original DNN model into believing it is a gibbon.

To mitigate the aforementioned kind of attack, solutions~\cite{gu2014towards,huang2015learning, miyato2015distributional} proposed generally follow the basic idea of {\em adversarial training} in which a DNN is trained with both samples from the original data distribution as well as artificially synthesized adversarial ones. A recent unification of previous approaches \cite{ororbia_ii_unifying_2016} showed that they were all special cases of a general, regularized objective function,\emph{DataGrad}.  While such a framework vastly improves the DNN's robustness to adversarial samples, the final model is still not completely resilient given that the adversarial sample space is unbounded. Hence, a newly trained DNN would only be robust with respect to previously observed adversarial samples (or for those relatively near to training samples of the underlying manifold if one uses the general approximation to \emph{DataGrad} developed in \cite{ororbia_ii_unifying_2016}).

Here, we present a new defense framework that increases the difficulty for attackers by crafting adversarial samples, i.e. making DNN models resilient to any adversarial sample whether or not they have been previously observed by DNN models. The basic idea underlying our new framework is to transform the input data into a new representation before presenting it to a DNN model. More specifically, the transformation employs a non-invertible dimensionality reduction approach that projects an input sample into a low dimension representation. As mentioned earlier, an attacker needs to perform an invertible computation to examine the feature/class importance through an auxiliary DNN model. With this
transformation, the complexity of such an invertible computation is significantly increased and the attacker is no longer able to perform feature/class examination.

Technically speaking, our framework is similar to a distillation process, which trains a DNN model using the knowledge transferred from a different model~\cite{papernot2015distillation}. However, it is fundamentally different from the distillation approach. In fact, the distillation approach does not improve the robustness of a DNN because it does not increase the complexity of the invertible computation. As we will demonstrate in Section~\ref{sec:motivation}, an attacker can easily construct an auxiliary model and craft adversarial samples even when the DNN model is trained through a distillation procedure.

In summary, this work makes the following contributions.

\begin{itemize}
  \item We conduct a detailed analysis on various existing defense mechanisms against adversarial samples, and demonstrate their limitations.
  \item We present a comprehensive framework that makes a DNN model resilient to adversarial samples by integrating an input transformation into the DNN model.
  \item We develop two new defense mechanisms by injecting different dimensional reduction methods into the proposed framework.
  \item We theoretically and empirically evaluate the DNN models, showing that our new defense framework is resilient to adversarial samples.
\end{itemize}

The rest of this paper is organized as follows. Section~\ref{sec:background} introduces the background of DNNs. Section~\ref{sec:motivation} discusses the limitations of existing defense mechanisms. Section~\ref{sec:framework} presents a unifying framework and Section~\ref{sec:DRmethods} develops two defense mechanisms using dimensionality reduction methods, both linear and nonlinear. In Section~\ref{sec:eval}, we evaluate our proposed framework. Section~\ref{sec:rw} discusses some related work followed by the conclusion in Section~\ref{sec:conclusion}.
\section{Background}
\label{sec:background}
We briefly introduce the well-established deep neural network (DNN) model and then describe how to generate adversarial samples from it in order to exploit its inherent flaws. Finally, we discuss our threat model.

\subsection{Deep Neural Networks}

\comment{
DNNs have been widely adopted for solving both supervised and unsupervised learning problems~\cite{lecun2015deep}. Here we focus on supervised architectures, since machine learning tasks targeting malicious behaviour in computer security have been mostly studied by analyzing well established malware data sets using supervised learning~\cite{pascanu2015malware}.
}

\begin{figure}[t]
\begin{center}
  \includegraphics[scale=0.70]{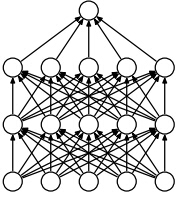}\\
  \caption{Standard neural network~\cite{srivastava2014dropout}  with two hidden layers, where neurons are fully connected. This architecture is often referred to as a feed-forward neural network~\cite{Goodfellow-et-al-2016-Book}}
  \label{fig:dnn}
\end{center}
\end{figure}

A typical DNN architecture consists of multiple successive layers of processing elements, or so-called ``neurons''. Each processing layer can be viewed as learning a different, more abstract representation of the original multidimensional input distribution. As a whole, a DNN can be viewed as a highly complex function that is capable of mapping original high-dimensional data points to a lower dimensional space in any nonlinear fashion. As shown in Fig~\ref{fig:dnn}, a typical DNN contains an input layer, multiple hidden layers, and an output layer. The input layer takes in each data sample in the form of a multidimensional vector. Starting from the input, computing the activations of each subsequent layer simply requires, at minimum, a matrix multiplication (where a weight/parameter vector, with length equal to the number of hidden units in the target layer, is assigned to each unit of the layer below) followed by summation with a bias vector. This process roughly models the process of a layer of neurons integrating the information received from the layer below (i.e., computing a pre-activation) before applying an elementwise activation function\footnote{There are many types of activations to choose from, including the hyperbolic tangent, the logistic sigmoid, or the linear rectified function, etc~\cite{Goodfellow-et-al-2016-Book}}. This integrate-then-fire process is repeatedly subsequently for each layer until the last layer is reached. The last layer, or output, is generally interpreted as the model's predictions for some given input data, and is often designed compute a parameterized posterior distribution using the softmax function (also known as multi-class regression or maximum entropy). This bottom-up propagation of information is also referred to as \emph{feed-forward} inference~\cite{hinton2007learning}.

During the learning phase of the model, the DNN's predictions are evaluated by comparing them with known target labels associated with the training samples. Specifically, both predictions and labels are taken as the input to a selected cost function, such as cross-entropy. The DNN's parameters are then optimized with respect to the cost function using the method of steepest gradient descent, minimizing prediction errors on the training set. Parameter gradients are calculated using back-propagation of errors~\cite{rumelhart1988learning}. Since the gradients of the weights represent their influence on the final cost, there have been multiple algorithms developed for finding optimal weights more accurately and efficiently~\cite{ngiam2011optimization}. 

\subsection{The Adversarial Sample Problem}
\label{adv_samples}
Even though a well trained model is capable of recognizing out-of-sample patterns, a deep neural architecture can be easily fooled by introducing perturbations to the input variables that are often indistinguishable to the human eye~\cite{42503}. These so-called ``blind spots'', or adversarial samples, exist because the input space of DNN is unbounded~\cite{Goodfellow14}. Based on this fundamental flaw, we can uncover specific data samples in the input space able to bypass DNN models. More specifically, it was studied in \cite{Goodfellow14} that attackers can find the most powerful blind spots via effective optimization procedures. In multi-class classification tasks, such adversarial samples can cause a DNN model to classify a data point into a random class besides the correct one (sometimes not even a reasonable alternative).

Furthermore, according to~\cite{42503}, DNN models that share the same design goal, for example recognizing the same image set, all approximate a common highly complex, nonlinear function. Therefore, a relatively large fraction of adversarial examples generated from one trained DNN will be misclassified by the other DNN models trained from the same data set but with different hyper-parameters. Therefore, given a target DNN, we refer to adversarial samples that are generated from other different DNN models but still maintain their attack efficacy against the target as \emph{cross-model adversarial samples}. 

Adversarial samples are generated by computing the derivative of the cost function with respect to the network's input variables. The gradient of any input sample represents a direction vector in the high dimensional input space. Along this direction, an small change of this input sample will cause a DNN to generate a completely different prediction result. This particular direction is important since it represents the most effective way of compromising the optimization of the cost function. Discovering this particular direction is realized by passing the layer gradients from output layer to input layer via back-propagation. The gradient at the input may then be applied to the input sample(s) to craft an adversarial example(s).

To be more specific, let us define a cost function $\mathcal{L}(\theta,X,Y)$, where $\theta$, $X$ and $Y$ denotes the weights of the DNN, the input data set, and the corresponding labels respectively. In general, adversarial samples can be crafted by adding to legitimate ones via an \emph{adversarial perturbation} $\delta x$. In~\cite{Goodfellow14}, the efficient and straightforward \emph{fast gradient sign} method was proposed for calculating adversarial perturbation as shown in in~\eqref{eq:delta_X}:
\begin{equation}
  \begin{aligned}
  \label{eq:delta_X}
	\delta x = \phi sign(\mathcal{J}_{\mathcal{L}}(\theta,x,y)),
  \end{aligned}
\end{equation}
here $\delta x$ is calculated by multiplying the sign of the gradients of legitimate samples with some coefficient $\phi$. $\mathcal{J}_{\mathcal{L}}$ denotes the gradients of the loss function $\mathcal{L}(\cdot)$ with respect to the data $x$. $y$ is the corresponding label of $x$. $\phi$ controls the scale of the gradients to be added. 

Adversarial perturbation indicates the actual direction vector to be added to legitimate samples. This vector drives a legitimate sample $x$ towards a direction that the cost function $\mathcal{L}(\cdot)$ is significantly sensitive to. However, it should be noted that $\delta x$ must be maintained within a small scale. Otherwise adding $\delta x$ will cause a significant distortion to a legitimate sample, leaving the manipulation to be easily detected. In the next section, we will develop two threat models that utilize the properties of adversarial samples.

\subsection{Threat Model}
The threat models introduced in this part all follow the line of exploiting the sensitivity of DNN models with respect to input samples. We first introduce an attack which utilizes direct knowledge of the target DNN model. This attack is only valid under the assumption that all detailed DNN information, regarding both architecture and exact parameters, is disclosed to an adversary. Since this assumption might be too strong to be practical in real world scenarios, we further present another threat model that is also effective for fooling a normal DNN but not restricted by the aforementioned assumption. 

\subsubsection{White Box Threat Model}
As previously introduced, an adversary can manipulate data samples to mislead a normal DNN by using gradients with respect to a loss function. In order to obtain the gradient of a data sample, the adversary needs to have access to the exact form of the cost function, the architecture of the DNN mode, as well as the tuned coefficients found as a result of the model's training procedure. Assuming that all of this information is available, an adversary can tweak any testing data sample towards a direction using its gradient information. 

To be more specific, the gradient of a test sample indicates the direction along which even small changes of this test sample will yield a significant difference in the cost function. Once the gradients are calculated, one may use the procedure detailed in Section \ref{adv_samples} to manipulate samples to trick the DNN into generating worst case prediction results. However, in the real world, detailed DNN information can be well protected using various data integrity techniques making the White Box assumption too strong. Therefore, we introduce another threat model not restricted by this assumption and is yet effective in attacking a normal neural architecture.

\subsubsection{Black Box Threat Model}
Since DNNs have become a well established machine learning method, the training procedures and typical cost functions used are common public knowledge. Furthermore, recall that a DNN is designed to approximate a highly nonlinear function that is capable of mapping original data samples into a space that makes prediction simpler. If we assume that a given training data and application scenario, then it follows that different DNN models are all tuned to behave similarly to one other. Therefore, an attacker could utilize a well-known learning algorithm, a typical architecture, and a similar data set to build an auxiliary model that accurately approximates the target model. Since the auxiliary model is in the adversary's full possession, he can easily generate cross-model adversarial samples as previously introduced. Therefore, a successful attack (using cross-model adversarial samples) can be conducted even most of the information about a target DNN is unavailable.

Based on these two threat models, our proposed DNN architecture is designed to be robust in both scenarios. Particularly, we provide both a theoretical proof and empirical demonstration of the robustness of our proposed DNN architecture even under the white box threat. In the following section, we will provide a thorough analysis of existing solutions for defending against adversarial samples. We will then show that the fundamental flaw introduced in this section is not fully and properly addressed by these solutions. 

\section{Motivation}
\label{sec:motivation}
We start this section with an overview of recent solutions developed for defending against adversarial samples. These solutions fall into two categories: 1) augmenting the training set and 2) enhancing model complexity. The former is mainly represented by adversarial training while the latter mainly combines various data transformations with a standard DNN. 

\subsection{Data Augmentation}
For image recognition, current publicly available image sets can be easily used as a starting point. Adversarial samples shown in Fig.~\ref{fig:panda} are blind spots specifically created to trick a DNN into misclassifying with high confidence. To resolve the blind spot issue, there have been many data augmentation methods proposed for deep learning tasks~\cite{Goodfellow14, ororbia_ii_unifying_2016}.
In principle, these methods expand their training set by combining known samples with potential blind spots. The same mechanism has also been employed for defending against adversarial samples, also known as adversarial training~\cite{ororbia_ii_unifying_2016}.
Here, we analyze the limitations of data augmentation mechanisms and argue that these limitations also apply to adversarial training methods.

Given the high dimensionality of data distributions that DNNs typically learn from, the input space is generally considered infinite~\cite{Goodfellow14}. This implies that there could also exist an infinite amount of blind spots, which are adversarial samples specific to DNN models. Therefore, data augmentation based approaches have the challenge of covering these very large spaces. Since adversarial training is a form of data augmentation, this approach is inherently limited in its ability to protect a DNN from blind spots.

Adversarial training can be formally described as adding a regularization term known as \emph{DataGrad} to a DNN's training cost function~\cite{ororbia_ii_unifying_2016}. The regularization penalizes the directions uncovered by adversarial perturbations (introduced in Section \ref{sec:background}). Therefore, adversarial training works to improve the worst case performance of DNN. Treating the DNN much like a generative model, adversarial samples are produced via back-propagation and mixed into the training set and directly integrated into the model's learning phase. 

Despite the fact that there exists infinite adversarial samples, adversarial training is effective for defending against those which are powerful and easily crafted. This is due to the fact that, in most adversarial training approaches~\cite{ororbia_ii_unifying_2016, Goodfellow14}, adversarial samples can be generated efficiently for a particular type of DNN. The fast gradient sign method~\cite{Goodfellow14} can generate a large pool of adversarial samples quickly while \emph{DataGrad}~\cite{ororbia_ii_unifying_2016} focuses on dynamically generating them per every parameter update. However, the simplicity and efficiency of generating adversarial samples also makes adversarial training vulnerable when these two properties are exploited in order to attack the adversarial training \emph{itself}. Given that there exist infinite adversarial samples, we would need to repeat an adversarial training procedure each time a new adversarial example is found. DataGrad, as mentioned \cite{ororbia_ii_unifying_2016}, could be viewed as taking advantage of adversarial perturbations to better explore the underlying data manifold--however, while this leads to improved generalization, it does not offer any guarantees in covering all possible blind-spots. More importantly, adversarial training does nothing to change the actual reversible nature of a DNN's architecture itself, which we argue is the more direct and ultimately more effective way in building a strong defense.

\subsection{Enhancing Model Complexity}
DNN models are already complex, with respect to both the nonlinear function they try to approximate as well as their layered composition of many parameters. The architecture is straightforward in order to facilitate the flow of information forwards and backward, which greatly alleviates the effort in generating adversarial samples. Therefore, several ideas~\cite{papernot2015distillation, gu2014towards} have been proposed to enhance the complexity of DNN architecture, aiming to improve the tolerance of complex DNN models with respect to adversarial samples generated from simple DNN models. 

\cite{papernot2015distillation} develops a \emph{defensive distillation} mechanism, which trains a DNN from data samples that are distilled by another DNN. By using the knowledge transferred from the other DNN, the learned DNN classifiers become less sensitive to adversarial samples. However, 
although shown to be effective, this method still relies on gradient flow from output to input. This is because both DNN models used in this architecture can be approximated by an adversary via training two other DNN models that share the same functionality and have similar performance. Once the two approximating DNN models are obtained, the adversary can generate adversarial samples specific to this distillation-enhanced DNN model. \cite{gu2014towards} proposed stacking an auto-encoder together with a normal DNN, similar to~\cite{papernot2015distillation}.  It was first shown that this auto-encoding enhancement increases DNN resilience to adversarial samples. However, the authors further demonstrate that this stacked model can be simply taken as a new DNN and easily generate new adversarial samples. 

Though the above approaches, both from data-augmentation and model complexity perspectives, have proven effective in handling samples generated from normal adversarial DNN models, they do not handle all adversarial samples. In light of this, we propose a framework that blocks the gradient flow from the output to input variables, a solution that prove effective even when the architecture and parameters of a given a DNN are publicly disclosed.
\section{Data Transformation Enhanced DNN Framework}
\label{sec:framework}
In this section, we shall fully specify our framework's design goals and choose a particular type of data transformation that will fulfill these goals.

\subsection{Design Goals}
\begin{figure*}[t]
\begin{center}
  \includegraphics[scale=0.70]{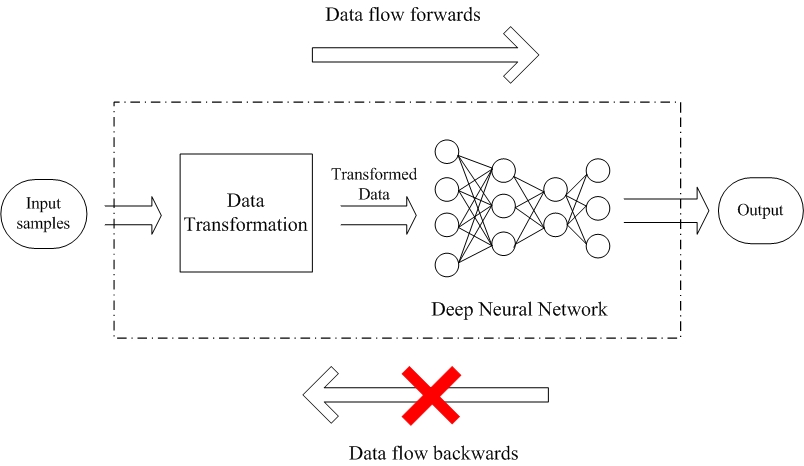}\\
  \caption{A data transformation enhanced DNN. In the feed-forward pass, input data is first projected to a particular space and the transformed data is then used to train a standard DNN classifier. However, the gradient flow is blocked by the data transformation.}
  \label{fig:frame}
\end{center}
\end{figure*}

Recall that many previously proposed solutions, especially adversarial training methods \cite{Goodfellow14, ororbia_ii_unifying_2016}, can be classified as forms of data augmentation. By their very nature, these approaches cannot possibly hope to cover the entire data space, unless perhaps they had access to all important representative points of the underlying manifold (which is highly unlikely in practice). This implies that attackers can always find adversarial samples targeted to any particular DNN model. More importantly, developing adversarial training methods has required the invention of efficient methods for generating adversarial samples \cite{Goodfellow14, 42503}, consequently providing more useful tools for attackers. 

This further facilitates new attacks specifically designed for DNN models learned using adversarial training algorithms. 

We argue that a robust DNN architecture has the property that adversarial samples cannot be generated from itself. This we consider to mean that the flow of error gradients from the output to input variables must be obstructed (while minimally affecting model generalization performance). Since most DNN models are simple and straightforward, their architectures are usually well known and hence easy to approximate. Therefore, a robust DNN satisfying our desired property can still be secure even under the most extreme circumstances, such as when all of its detailed parameters are disclosed to adversaries. Our framework specifically must satisfy two goals:
\begin{itemize}
  \item It has minimal impact on the performance of a DNN model when legitimate samples are seen.
  \item It is resilient to cross-model adversarial samples as introduced in Section \ref{sec:background}.
  \item It `\emph{blocks} gradient flow from reaching the input. 
\end{itemize}

\subsection{Framework Overview}
Adversarial samples are generated via backpropagation during training, as described in Section ~\ref{sec:background}. This implies that by following the path of gradient flow, adversarial samples can be generated quite conveniently. In order to prevent this, we propose blocking this gradient path by combining the DNN architecture with particular data transformation methods. Specifically, we insert a data transformation module before the input layer of the DNN, in effect preforming a special preprocessing of input samples. As a result, the procedure for generating adversarial samples will only work for the transformed data, if the right kind input transformation is chosen such that information cannot reach the original observed variables directly from objective function. Furthermore, even when an adversary successfully generates adversarial samples based on the transformed data, we can \emph{guarantee} that this adversary will be unable to map the adversarial data samples to the original input space if the data transformation employed is \emph{non-invertible}. More formally, we define a data transformation method as non-invertible when it satisfies either of the following properties: (1)~inverting the data transformation is computationally too complex to be tractable; or (2)~inverting the data transformation will cause significant reconstruction error.

Our proposed framework, graphically depicted in Fig~\ref{fig:frame}, has several properties that guarantee that a DNN is well protected against adversarial samples while suffering at most only trivial changes in performance. First, any non-invertible data transformation method can ensure that the gradient flow in the framework is effectively blocked. This critically ensures that an adversary cannot generate samples using our framework using a method like the fast gradient sign method (see Section~\ref{sec:background}). Second, the framework is capable of legitimately handling test samples since the DNN model is trained on top of a data transformation that preserves crucial information contained in the original input distribution. Third, any non-invertible data transformation method may be used in tandem with a normal DNN model under our unified framework. 

However, it is important to note that integrating a DNN with a data transformation may be computationally expensive, which might make this defense less attractive given that standard DNN training is already computationally intensive. We resolve this issue by exploring data transforms that are computationally efficient and more importantly, incremental. The latter requirement is essential given that any data transformation method must be capable of handling unseen samples as they are presented. Otherwise, the data transformation will need to be retrained, and subsequently, the DNN on top of the newly retrained transform layer.

Dimensionality reduction is one particular data transformation mechanism that satisfies these design objectives. First, dimensionality reduction methods are often designed to preserve at least the most important aspects of the original data. Second, dimensionality reduction can serve as a filter for adversarial perturbations when a DNN is confronted with cross-model adversarial samples. Third, dimensionality reduction helps reduce the dimensionality of the input distribution that is fed into the DNN. Finally, it is easier to develop non-invertible data transformation methods, since recovering higher dimensional data from lower dimensional data is difficult. The following sections introduce details of two developed defence mechanisms using different non-invertible dimensional reduction methods.

\section{Data transformation enhanced DNNs}
\label{sec:DRmethods}
We present two variations of an adversary resilient DNN architecture that satisfies the two key design goals of our proposed defense framework (Section ~\ref{sec:framework}). In particular, one variant utilizes a linear method while the other makes use of a non-linear approach. With respect to the linear method, we show that, theoretically, there is a lower bound on the reconstruction error. More importantly, we design our linear method to have a reconstruction error that is significantly larger than this lower bound which in turn satisfies the characteristics we defined for a non-invertible data transformation. With respect to the second approach, we employ a nonlinear dimension reduction approach that we prove satisfies the second characteristic of a non-invertible data transformation. Critically, we show that, in order to recover the original data from from this low dimensional representation, an inversion of the corresponding dimensional reduction method is required, which can be converted to a quadratic problem with a non-positive, semi-definite constraint. This belongs to a class of NP-hard problems, according to~\cite{d2003relaxations}.  
\subsection{Designed Linear Mapping (DLM) DNN}
We first propose a novel linear dimensional reduction method, which stems from principal component analysis (PCA). Then, a straightforward strategy is introduced for reconstructing the original high dimensional data by inverting our proposed transformation. In addition, we provide a theorem that places the lower bound on the reconstruction error. The inversion strategy will be further used in Section~\ref{sec:eval} to demonstrate that our proposed method satisfies the requirements of non-invertibility.

PCA is one of the most widely adopted dimensional reduction methods, especially since it is computationally efficient and easy to implement~\cite{jolliffe2002principal}. These are useful properties when considering PCA as a possible transform to combine with a DNN, since the DNN itself is already computationally expensive. Additionally, PCA preserves critical information by finding a low-dimensional subspace with maximal variance. However, PCA can be easily inverted to reconstruct original data samples given their lower dimensional mappings. Moreover, the reconstructed data usually exhibits insignificant distortion when compared to the original data (see Figure~\ref{fig:inverted_PCA} (a)). This implies that, if an attacker generates adversarial samples from a DNN trained on low dimensional mappings, it would be simple to generate high dimension adversarial samples by mapping this low dimensional data back to the original dimension. We will show in Section \ref{sec:eval} the effectiveness of inverting PCA by examining several reconstructed images from the MNIST data set.

PCA preserves meaningful features of the original data when mapping them to a lower dimension. Given a data matrix $X\in \mathbb{R}^{n\times p}$, the transformation matrix $W$ can be obtained by solving the following optimization function:
\begin{equation}
  \begin{aligned}
  \label{eq:pca}
	\textup{arg}\;\underset{Y,W}{\textup{min}} \frac{1}{2}\left | \left | X-YW^{T} \right | \right |_{F}
  \end{aligned}
\end{equation}
where $W \in \mathbb{R}^{p \times q}$, $W^{T} W =I_{q}$ and $Y \in \mathbb{R}^{n \times q}$. According to the Eckart-Young Theorem~\cite{eckart1936approximation}, the optimal solution is obtained when $W$ consists of the $q$ largest eigenvalues of $X ^{T}X$. Therefore, the low dimensional mappings can be computed as follows:
\begin{equation}
  \begin{aligned}
  \label{eq:linearpca}
	Y=XW
  \end{aligned}
\end{equation}
Accordingly, we can approximately reconstruct the high dimensional $X$ from the transformed data $Y$ according via:
\begin{equation}
  \begin{aligned}
  \label{eq:repca}
	\hat{X} = YW^{T}
  \end{aligned}
\end{equation}
which represents the process of reconstructing high dimensional approximation using only low dimensional mappings and a transform matrix. Note that this process has low computational cost--we only need to calculate the inverse transformation of $Y$. In Section~\ref{sec:eval}, we show that the reconstructed data is quite similar to original data. Therefore, using PCA alone for a data transformation does not satisfy the non-invertible criteria we introduced in Section~\ref{sec:framework}, and thus insufficient for crafting an adversary-resilient DNN.

To deal with this problem and yet preserve computational efficiency, we equip PCA with our first non-invertible characteristic. To do this, we propose a novel dimension reduction method we call a designed linear mapping (DLM). DLM is designed to combine the lower dimensional mapping generated by PCA with other lower dimensional mappings generated by multiplying the original data with a column-wise, highly correlated transformation matrix. This design ensures that the PCA operation continues to preserve the critical information while the column-wise highly correlated transformation matrix guarantees that inverting the DLM will generate significant reconstruction error. To explain the consequence of this, we now introduce DLM in detail and examine its properties.

Much as in~\eqref{eq:linearpca}, we shall formally define DLM to be:
\begin{equation}
  \begin{aligned}
  \label{eq:drA}
	Y = X C^{T} + \omega,
  \end{aligned}
\end{equation}
where  $X \in \mathbb{R}^{n \times p}$ the same as introduced earlier, except that $Y \in \mathbb{R}^{n \times p_{c}}$. $\omega \in \mathbb{R}^{n \times p_{c}}$ denotes a normally distributed noise matrix, where each entry of $\omega$ generated from a normal distribution $N(0, \sigma^{2})$. $C \in \mathbb{R}^{p_c \times p} $ is the transformation matrix obtained by following equation:
\begin{equation}
  \begin{aligned}
  \label{eq:C}
	C = [B;A],
  \end{aligned}
\end{equation}
where C is constructed by combining a loading matrix $B \in \mathbb{R}^{p_b \times p}$ obtained via PCA with a designed matrix $A \in \mathbb{R}^{(p_c - p_b) \times p}$, of which all columns are highly correlated. This combination integrates PCA's information-preserving effects into our DLM. As such, the lower dimensional projection $Y$ can provide a better representation of the original $X$.

Since the DLM described by~\eqref{eq:drA} has a simple linear form, we estimate reconstruction $\hat{X}$ for $X$ using high-dimensional linear regression~\cite{raskutti2011minimax} (we omit calculation details due to space constraints). According to Theorem 1 in~\cite{raskutti2011minimax}, under certain assumption, we can obtain a lower bound of the reconstruction error, which is the $L_{2}$ norm of the difference between $X$ and $\hat{X}$ as shown in~\eqref{eq:theorem1}:
\begin{equation}
  \begin{aligned}
  \label{eq:theorem1}
	\left( L_{2}(X, \hat{X}) \right )^2 \geq \kappa_0 \mbox{ } \sigma^{2} \frac{s \mbox{ } log(p/s)}{p_{c}},
  \end{aligned}
\end{equation}
where $s$ denotes the sparsity of $X$. $\kappa_0$ is a constant whose value depends closely on the data set. Therefore, given a certain set of data, any linear transformation method is restricted by a constant lower bound calculated according to~\eqref{eq:theorem1}. In addition, according to Theorem 2 in~\cite{raskutti2011minimax}, there also exists an upper bound of the reconstruction error as follows:
\begin{equation}
  \begin{aligned}
  \label{eq:theorem2}
	\left( L_{2}(X, \hat{X}) \right )^2 \leq f(C) \frac{s \mbox{ } log(p)}{p_{c}},
  \end{aligned}
\end{equation}
where $f(C)$ is a function of $C$. According to~\cite{raskutti2011minimax}, the upper bound of the reconstruction error depends on both the data transformation matrix $C$ and noise $\omega$. When $C$ is a an independent correlation matrix, as in PCA, then the upper bound will approach the aforementioned lower bound hence restricting the reconstruction error within a narrow range close to the lower bound. However, since we specifically design $C$ to be highly correlated, the upper bound will be significantly larger than the lower bound~\cite{fornasier2011compressive,Compressed2006Donoho}, and thus result in a larger range for the reconstruction error. 
% AO: cut out last bit b/c it was redundant
%To validate our design, in following Section \ref{sec:eval}, we will empirically demonstrate that the reconstruction errors obtained by inverting DLM as introduced above.

\subsection{Dimensionality Reduction by Learning an Invariant Mapping (DrLIM) DNN}
As introduced in Section~\ref{sec:background}, adversarial samples are generated by changing legitimate samples with small perturbations. But when processed by normal DNN models, the decisions made in a lower dimensional space are completely different from those made for legitimate samples, even though adversarial samples are highly similar to legitimate ones. Note that this characteristic also occurs in cross-model adversarial samples. Therefore, we intend to employ a dimensionality reduction method that preserves the similarity of high dimensional samples in their lower dimensional mappings. Furthermore, our method needs to be capable of extracting critical information contained in the original data. Since the training of a DNN is already computationally intensive, our approach needs to be incremental in order to avoid the need for retraining the DNN. 

Because of these considerations, we employ the dimensionality reduction method \emph{DrLIM} proposed in \cite{Hadsell2006Dimensionality}. DrLIM is specifically designed for preserving similarity between pairs of high dimensional samples when they are mapped to a lower dimensional space. The method restricts the lower dimensional mapping of cross-model adversarial samples to a vicinity where it is filled by mappings of legitimate samples that are highly similar to these cross-model adversarial samples. As a result, there is a significantly lower chance that an adversarial sample acts as an outlier in the lower dimensional space, since its mapped location is bounded by the mapped locations of similar, legitimate samples. DrLIM can also be used in an online setting. 

More importantly, we theoretically prove that inverting DrLIM is an NP-hard problem. Therefore, DrLIM is suitable for our framework in that it satisfies the second characteristic of non-invertiblity defined in Section \ref{sec:framework}. But first, we briefly review DrLIM (for a detailed explication please see~\cite{Hadsell2006Dimensionality}).

DrLIM consists of a convolutional neural network (CNN) model designed for optimizing the cost function:
\begin{equation}
  \begin{aligned}
  \label{eq:cost}
	\sum_{i=1}^{P} L\left (W, (Y, X_{i_1}, X_{i_2})^i\right ),
  \end{aligned}
\end{equation}
where $W$ denotes the coefficients. $X_{i_1}$ and $X_{i_2}$ denote the $i$th pair of input sample vectors with $i = 1\dots P$. $Y$ is a binary label assigned to each pair of samples, with $Y=0$ denoting a similar pair of $X_{i_1}$ and $X_{i_2}$, and $Y=1$ for dissimilar pairs. Note that the similarity of each pair is not limited to any particular distance-based measure. This means that any prior knowledge can be utilized in representing similarity, including manually assigned similarity and dissimilarity. Thus, the classical approach for measuring a Euclidean distance between samples for representing dissimilarity can be enhanced with prior knowledge. Let the loss function for measuring the cost for each pair of samples be defined as:
\begin{equation}
  \begin{aligned}
  \label{eq:loss}
	L\left(W, Y, X_{1}, X_{2}\right ) =& (1-Y)\frac{1}{2}\big(D(X_{1}, X_{2})\big)^2 + \\
	&\frac{Y}{2}\{max\big(0, m-D(X_{1}, X_{2})\big)\}^2,
  \end{aligned}
\end{equation}
where $D(X_{1}, X_{2}) = \left \| G(X_{1}) - G(X_{2})\right \|_2$ is the Euclidean distance measured between the output lower dimension mapping $G(X_{1})$ and $G(X_{1})$ for the sample pair $X_{1}$ and $X_{2}$. Let $m$ be a predefined constant which indicates whether all dissimilar pairs are pushed or pulled towards to maintain a constant distance $m$ ($i$ omitted for simplicity).

Since $G$ represents a mapping by the CNN to enable the recovery of high dimensional data from the low dimensional data $G(X)$, we need to first get $G^{-1}(X)$. For the forward pass of a conventional neural network, it is not guaranteed that the weight matrices are invertible~\cite{zeiler2014visualizing}, implying that information lost during pooling cannot be recovered. Thus, it is very difficult to compute $G^{-1}(X)$ and recover the original data from a low dimensional representation. Since inverting the CNN is nearly impossible, one option is to reconstruct original $X$ according to~\eqref{eq:loss} given $W$ and $Y$. In the following, we demonstrate that even this approach can be mapped to a NP-hard problem.

As discussed before, the most important property of DrLIM that allows it to fit into our framework is that it is provably non-invertible. Assuming $G(X)$ takes a simple linear form of  $G(X) = WX$, then we have $D(X_1, X_2)^2 = (X_1-X_2)^T W^T W(X_1-X_2)$. Here we denote $\delta X = (X_1-X_2)$. Following this assumption, we can reformulate~\eqref{eq:loss} as follows:
\begin{equation}
\begin{aligned}
& \underset{\delta X, z}{\text{min}}
& & \sum (1-Y) \delta X^T W^T W \delta X + Y z^2,  \\
& \text{s.t.}
& & z \ge 0, \\
& & &  z \ge m-\sqrt{\delta X^T W^T W \delta X},
%& & & z + (X_1-X_2)^T W^T W (X_1-X_2) \ge m.
\label{eq:qcqp}
\end{aligned}
\end{equation}
where $z = max\big(0, m-D(X_{1}, X_{2})\big)$. Here we reformulate the second constraint as $\sqrt{\delta X^T W^T W \delta X} \ge m - z$. Since $m - z \ge 0$, we have following:
\begin{equation}
\begin{aligned}
\delta X^T W^T W \delta X \geq (m-z)^2.
\label{eq:nonpos_sdp}
\end{aligned}
\end{equation}
Therefore, $W$ is positive semi-definite. When both sides of~\eqref{eq:nonpos_sdp} are multiplied by $-1$ and substituted into~\eqref{eq:qcqp}, we find that:
\begin{equation}
\begin{aligned}
& \underset{\delta X, z}{\text{min}}
& & \sum (1-Y)\delta X^T W^T W \delta X  + Y z^2,  \\
& \text{s.t.}
& & z \ge 0, \\
& & &  - \delta X^T W^T W \delta X \leq - (m - z)^2.
%& & & z + (X_1-X_2)^TWW^T(X_1-X_2) \ge m.
\label{eq:npqcqp}
\end{aligned}
\end{equation}
From earlier work~\cite{Vavasis1991Nonlinear}, the formulation~\eqref{eq:npqcqp} implies a quadratic problem with a non-positive semi-definite constraint, which is an NP-hard problem. 

Note that solving~\eqref{eq:npqcqp} can yields the distance $\delta X$. There are multiple pairs of $X_1$ and $X_2$ that satisfy that $\delta X = (X_1-X_2)$. This makes the problem even harder to solve. Additionally, since the linear relaxation~\eqref{eq:npqcqp} is already NP-hard, the original problem~\eqref{eq:loss} is also NP-hard given that $G(X)$ is commonly regarded as a nonlinear function approximated by a neural network.

\section{Evaluation}
\label{sec:eval}

\begin{figure*}[t]
    \centering
    \begin{tabular}{c}
        \includegraphics[width=0.60\textwidth]{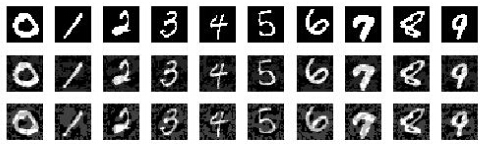}
    \end{tabular}
    \caption{Examples of different legitimate adversarial samples from MNIST. The top row shows examples of legitimate images, in comparison with the middle and bottom rows, in which adversarial samples generated from model $A$ and $A_{ADT}$ are shown, respectively. Clear distortion of images can be observed in the bottom two rows. Moreover, the distortion is increased for adversarial samples generated from $A_{ADT}$. This implies that model $A_{ADT}$ is more robust than model $A$, which can be attributed to adversarial training. More specifically, an attack on model $A_{ADT}$ requires yet further perturbation which makes it easier for humans to identify the assault.}
    \label{fig:multi_ad_train}
\end{figure*}

% AO: do we need this little intro? I say we just go right into the 1st sub-section
%In this section, we first demonstrate the limitation of adversarial training as previously introduced in Section \ref{sec:motivation}. Our demonstration is conducted by attacking a DNN model trained using different adversarial samples. Then we evaluate the classification performance of our proposed framework when confronted with both normal and adversarial samples. 

%Since in Section \ref{sec:DRmethods}, we have discussed about our framework for remaining secure even when all parameters are disclosed to an adversary. To examine this characteristic, we further assume that an adversary can indeed access all information of our framework. Therefore, we demonstrate the reconstruction errors for both proposed dimension reduction methods calculated following the procedure introduce in  Section \ref{sec:DRmethods}. At last, we visualize several examples of reconstructed samples, assuming the reconstruction errors are equal to the lower bound. In practice, the lower bound is considerably smaller than the reconstruction errors.

\subsection{Experiments Settings}
We evaluate our framework by performing experiments on the the widely-used MNIST data set~\cite{lecun1998mnist}. MNIST contains a training split with $600000$ greyscale images of handwritten digits and a test set containing $10000$ images. Each image has a dimensionality of $28 \times 28 = 784$ pixels.

In the following experiments, we evaluate the proposed approaches under two types of adversarial samples.  In order to first demonstrate that our mechanisms do indeed preserve the classification performance of the DNN, we test them with the original test set. We then test our methods with cross model adversarial samples to show that we achieve our secondary design goal. 

\subsection{Limitations of Adversarial Training}
Recently, adversarial training has been shown to be effective in decreasing the classification error rates at test time~\cite{Goodfellow14, ororbia_ii_unifying_2016}. Adversarial training samples are generated using the fast gradient sign method (Section \ref{sec:background}). At test time, new `\emph{adversarial test samples}' are generated based on unseen, legitimate test samples, given that the original DNN model is readily accessible. In this case, adversarial training results in improved robustness. 
%However, we further evaluate adversarial-trained DNNs against test samples generated from other normal DNN models. <-- redundant

We next build two different DNNs (model $A$ and $B$) that share the same purpose--image recognition. Furthermore, we utilize adversarial training in learning both models A and B, which we denote as models $A_{ADT}$ and $B_{ADT}$. Note that all following experiment results are the result of evaluating model $A_{ADT}$ using different testing samples. The results appear in Table~\ref{tab:ad_not_good}. The second row `\emph{Legitimate}' presents the classification error rates achieved by model $A_{ADT}$ when testing with normal samples.  In the next third and fourth row, we show classification error rates using adversarial samples generated from model $A$ and model $B$ respectively. While the classification error in both settings is lower than $30\%$, the error rate obtained when testing with adversarial samples generated from model $A$ itself is higher than the error rate found when testing with adversarial sample generated from a different model $B$. We speculate that this is possible because adversarial samples generated from a specific model might be more powerful for attacking that specific model. According to these results, adversarial training is indeed effective, depending on which external normal DNN model is used as the adversary. 

\begin{table}[b]
\centering
\caption{Classification performance of testing an adversarial training enhanced model with various adversarial samples}
\label{tab:ad_not_good}
\begin{tabular}{cc}
\hline
Different testing sets           & \begin{tabular}[c]{@{}c@{}}Classification error rates \\ of model $A_{ADT}$ \end{tabular} \\ \hline
Legitimate                                                        & 0.0213                                                                               \\ \hline
Adversarial testing samples from $A$             & 0.2506                                                                               \\ \hline
Adversarial testing samples from $B$             & 0.1633                                                                               \\ \hline
Adversarial testing samples from $A_{ADT}$ & 0.7810                                                                               \\ \hline
Adversarial testing samples from $B_{ADT}$ & 0.5715                                                                               \\ \hline
\end{tabular}
\end{table}

However, as previously mentioned in Section \ref{sec:motivation}, DNN models learned using adversarial training are still vulnerable to unseen adversarial samples crafted specifically to target them. To demonstrate this limitation, we show the results of testing model $A_{ADT}$ with new adversarial test samples generated from itself in the fifth row of Table~\ref{tab:ad_not_good}. The classification error rate is $78.10\%$, which is considerably higher than $25.06\%$ when testing model $A_{ADT}$ with adversarial testing samples generated from model $A$. This result is consistent with our previous analysis that the effectiveness of adversarial training is limited. In addition, we also present several visual samples of legitimate MNIST images as well as adversarial images generated from both model $A$ and $A_{ADT}$ in Fig.~\ref{fig:multi_ad_train}. As shown in Fig.~\ref{fig:multi_ad_train}, the legibility of image samples decreases as the attack power of these samples is increased. This indicates that an adversary would need to balance a trade-off between increasing attack strength and maintaining recognizable.

In the last row of Table~\ref{tab:ad_not_good}, we further show that the classification error rate of $57.15\%$ when testing model $A_{ADT}$ with adversarial samples generated from model $B_{ADT}$. Although the error rate is lower than $78.10\%$, it is still considerably higher than both $16.33\%$ and $25.06\%$ shown in the third and fourth rows. This result demonstrates that adversarial samples generated from enhanced DNN models maintain their cross-model efficacy. It is important to note that the effectiveness of the attack varies across models, an effect observed when using ordinary adversarial samples. 

\subsection{Classification Performance}
%In the following, we focus on the classification performance of the proposed DLM-DNN and DrLIM-DNN models. Specifically, performance is evaluated by testing both the DLM-DNN and DrLIM-DNN using legitimate samples and adversarial samples.  

\subsubsection{Classification Performance of DLM-DNN}
In this experiment, we fix the reduced dimensionality to $100$. These mappings are found by DLM and PCA. In order to better explore the effect of combining DLM with PCA, we vary the percentage $P_{pca}$ of PCA mappings used in the fixed $100$ dimension. Meanwhile, the percentage of DLM mappings used varies according to $100 - P_{pca}$. In addition, we also change the level of noise added to study its influence on classification performance.

We first show the classification performance when testing with legitimate samples in the column named as `\emph{Legitimate}' in Table~\ref{tab:DLM_DrLIM_DNN}. The noise coefficient is set to be either $0.1$ or $0.3$, while $P_{pca}$ varies from $5\%$ to $95\%$. We first notice that our proposed DLM-DNN does result in any significant performance degradation compared to adversarial training, especially when the noise coefficient is equal to $0.1$. However, it is noticeable that with a larger noise coefficient of $0.3$, the classification error rate of DLM-DNN goes up. This performance degradation is due to the increase of noise injected into the lower dimensional mappings. Therefore, we conclude that if properly set, DLM-DNN can result in performance comparable to adversarial training.
\begin{table}[t]
\centering
\caption{Classification performance of DLM-DNN and DrLIM-DNN}
\label{tab:DLM_DrLIM_DNN}
\begin{tabular}{ccccc}
\hline
\multicolumn{3}{c}{\multirow{2}{*}{\begin{tabular}[c]{@{}c@{}}Trained\\ Model\end{tabular}}}                                        & \multicolumn{2}{c}{\begin{tabular}[c]{@{}c@{}}Classification error rates\\ with different testing sets\end{tabular}} \\ \cline{4-5} 
\multicolumn{3}{c}{}                                                                                                                & Legitimate                                               & Adversarial                                               \\ \hline
\multicolumn{3}{c}{Normal DNN}                                                                                                      & 0.0198                                                   & 0.8981                                                    \\ \hline
\multicolumn{3}{c}{Adversarial Training Enhanced DNN}                                                                               & 0.0213                                                   & 0.2506                                                    \\ \hline
\multirow{10}{*}{DLM-DNN} & \multirow{5}{*}{\begin{tabular}[c]{@{}c@{}}Noise\\ coefficient of 0.1\end{tabular}} & PCA(95\%) & 0.0226                                                   & 0.3591                                                    \\ \cline{3-5} 
                          &                                                                                     & PCA(75\%) & 0.0247                                                   & 0.3211                                                    \\ \cline{3-5} 
                          &                                                                                     & PCA(50\%) & 0.0258                                                   & 0.2893                                                    \\ \cline{3-5} 
                          &                                                                                     & PCA(25\%) & 0.0268                                                   & 0.2735                                                    \\ \cline{3-5} 
                          &                                                                                     & PCA(5\%)  & 0.3101                                                   & 0.5212                                                    \\ \cline{2-5} 
                          & \multirow{5}{*}{\begin{tabular}[c]{@{}c@{}}Noise\\ coefficient of 0.3\end{tabular}} & PCA(95\%) & 0.0386                                                   & 0.2869                                                    \\ \cline{3-5} 
                          &                                                                                     & PCA(75\%) & 0.0403                                                   & 0.2685                                                    \\ \cline{3-5} 
                          &                                                                                     & PCA(50\%) & 0.0427                                                   & 0.2609                                                    \\ \cline{3-5} 
                          &                                                                                     & PCA(25\%)& 0.0452                                                   & 0.2699                                                    \\ \cline{3-5} 
                          &                                                                                     & PCA(5\%) & 0.3710                                                   & 0.5529                                                    \\ \hline
\multicolumn{3}{c}{DrLIM-DNN}                                                                                                       & 0.0384                                                   & 0.1380                                                    \\ \hline
\end{tabular}
\end{table}

\begin{figure}[b]
    \centering
    \begin{tabular}{c}
        \includegraphics[width=0.40\textwidth]{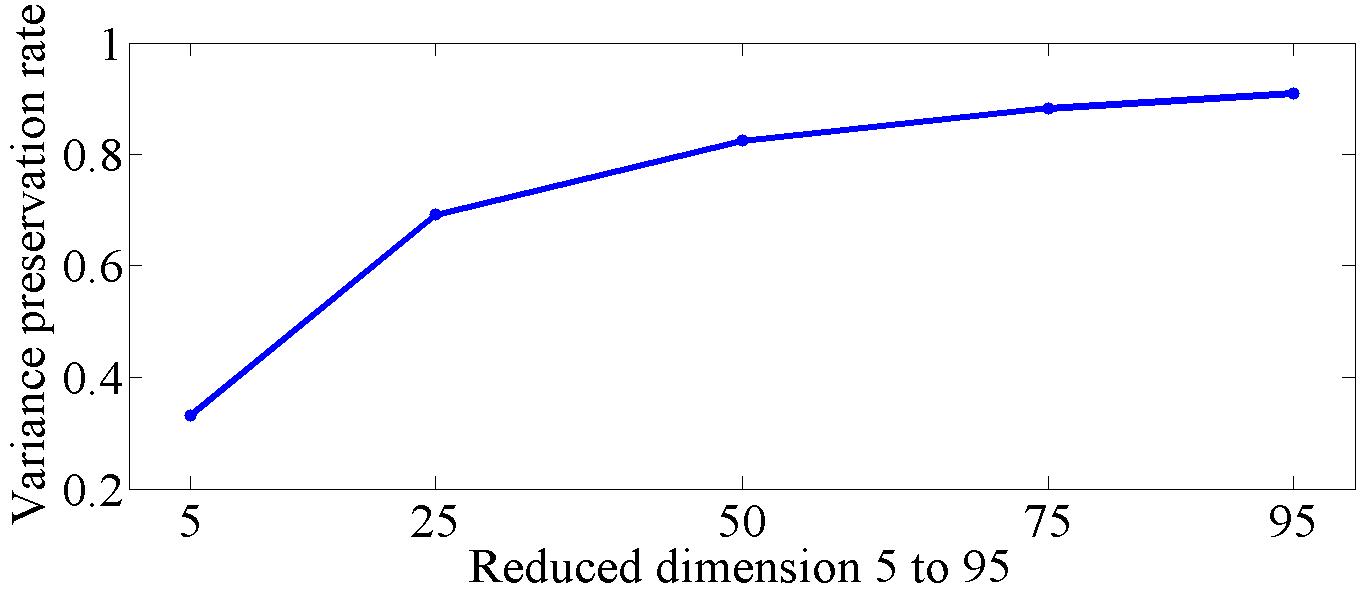}
    \end{tabular}
    \caption{Variance preservation rates with different reduced dimensions of PCA.}
    \label{fig:pcadev}
\end{figure}

We further examine the influence of varying $P_{pca}$ on classification performance. As shown in Table~\ref{tab:DLM_DrLIM_DNN}, the classification performance slightly improves in this case. This also happens when the noise coefficient is set to $0.3$. To better explain the performance boost, we further show the variance in preservation rates as a function of different dimensionalities using PCA in Fig.~\ref{fig:pcadev}. We show that PCA preserves over $90\%$ of the critical information within samples from the MNIST data set when mapping original data to the $100$ dimensional space. However, according to Fig.~\ref{fig:pcadev}, when $P_{pca} = 5\%$, information preserved by PCA is insufficient for representing the original samples well. Meanwhile, the $95\%$ mapping obtained from DLM is not as effective as PCA mappings for representing the original samples. Therefore, this combination leads to the highest classification error rates for both noise coefficient equal to $0.1$ and $0.3$. However, as $P_{pca}$ reaches $25\%$, enough information is preserved resulting in only a slight decrease in classification error. Meanwhile, when $P_{pca}$ varies from $25\%$ to $95\%$, the benefit of preserving any further information diminishes as with only a negligible decrease in the error rate. 

\begin{figure*}[t]
    \centering
    \begin{tabular}{cc}
        \includegraphics[width=0.40\textwidth]{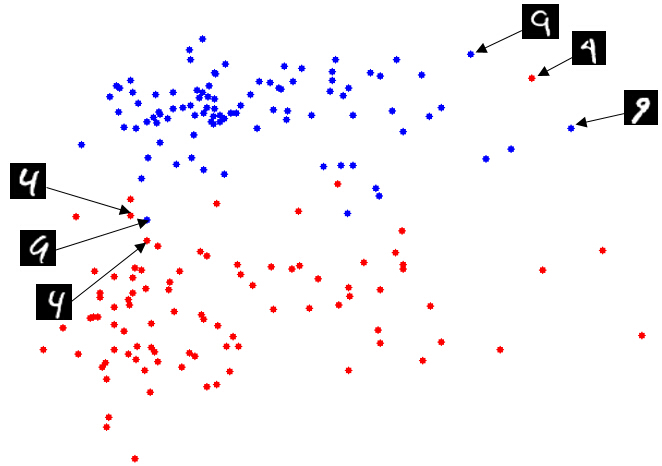} & \includegraphics[width=0.40\textwidth]{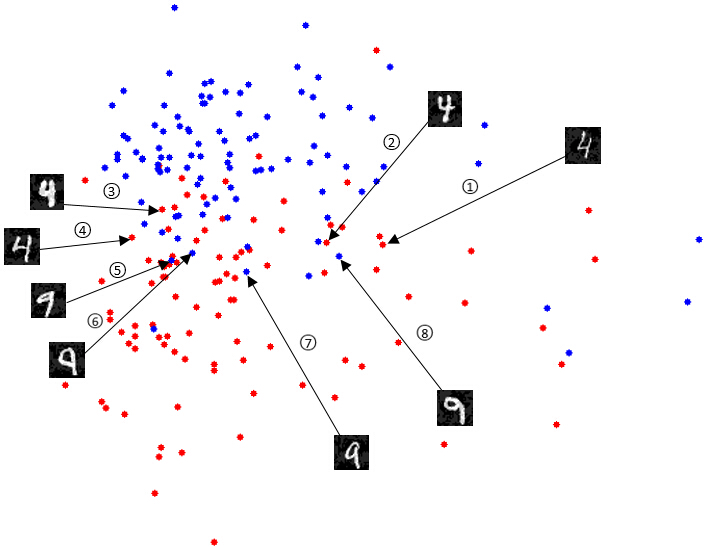} \\
        (a) 2D mapping of legitimate samples & (b) 2D mapping of adversarial samples \\[6pt]
    \end{tabular}
    \caption{2D mapping generated by DrLIM. In both (a) and (b), we show the 2D projection of 200 legitimate samples and 200 corresponding adversarial samples. Among these samples, half are randomly selected from class 4 (red dots) while the other half is randomly selected from class 9 (red dots). In (a), we highlight two outliers from both class 4 and 9 as well as their neighbours. Similarly, we highlight 8 outliers from both classes and mark each outlier sequentially by assigning each with an index, ranging from 1 to 8.}
    \label{fig:DrLIM}
\end{figure*}

We next evaluate the classification performance of DLM-DNN when confronted with adversarial samples. We list the classification error rates in the column noted as `\emph{Adversarial}'. According to Table~\ref{tab:DLM_DrLIM_DNN}, the error rates obtained by the DLM-DNN are considerably lower than that of a standard DNN, $0.8981$. Again, when $P_{pca}$ is properly set, the DLM-DNN achieves results comparable to adversarial training. The highest error rate is obtained when $P_{pca} = 5\%$, for either noise coefficient is set to $0.1$ or $0.3$. This is consistent from our previous analysis for testing with legitimate samples. Interesting enough, as $P_{pca}$ ranges from $25\%$ to $95\%$, classification error goes up. This observation might imply that the impact of adversarial samples is mitigated to a larger degree when more random disturbances are added. 

\subsubsection{DrLIM-DNN Classification Performance}
In this experiment we demonstrate the classification performance of DrLIM-DNN. The training set used for evaluation includes $5$ classes from the MNIST data, and each class contains $2000$ samples. For testing, each of the $5$ classes contains $1000$ testing samples. The adversarial testing samples are generated based on this testing set. 

For training the DrLIM, we label a pair of image samples as similar when they have the same label. This simplifies the training of DrLIM utilizing strong prior knowledge. We further set the reduced dimension to $30$ during the experiments. Classification performance is shown in the bottom part of Table~\ref{tab:DLM_DrLIM_DNN}. According to these results, using DrLIM-DNN results in a slightly higher error rate ($0.0384$) when testing with legitimate samples, but achieves a significant improvement in performance ($0.1380$) when testing adversarial samples. Especially in the latter case, DrLIM-DNN shows higher robustness when compared to adversarial training.

As previously introduced in Section~\ref{sec:DRmethods}, DrLIM is designed with the objective of preserving similarity between a pair of high dimensional samples when mapped to lower dimensional space. In order to better illustrate this property, we first map 200 legitimate examples of MNIST image samples into 2D dimension using DrLIM. According to the visualization of the 2D mapping shown in Fig.~\ref{fig:DrLIM}(a), DrLIM effectively preserves the similarity of legitimate samples in the mapped 2D space. We notice some outliers and hence highlight them and their neighbours by showing their corresponding images. For example, in Fig.~\ref{fig:DrLIM}(a), one of the outliers is a blue dot representing 9 surrounded by two red nots representing 4. Meanwhile the other outlier is a red 4 neighbouring to two blue 9s. As shown in Fig.~\ref{fig:DrLIM}(a), these outliers are hard to recognize, even for human.  

\begin{table}[b]
\centering
\caption{Classification confidence obtained from normal DNN and DrLIM-DNN}
\label{tab:DrLIMvsdnn}
\begin{tabular}{ccc}
\hline
\multirow{2}{*}{Outlier No.} & \multicolumn{2}{c}{\begin{tabular}[c]{@{}c@{}}Classification confidence \\ of testing adversarial samples\end{tabular}} \\ \cline{2-3}
                             & Normal DNN                                            & DrLIM-DNN                                            \\ \hline
1                            & 0.9995                                                  & 0.5196                                           \\ \hline
2                            & 0.9721                                                  & 0.5290                                           \\ \hline
3                            & 0.9989                                                  & 0.6220                                           \\ \hline
4                            & 0.9921                                                  & 0.5646                                           \\ \hline
5                            & 0.9998                                                  & 0.5903                                           \\ \hline
6                            & 0.9997                                                  & 0.5402                                           \\ \hline
7                            & 0.9998                                                  & 0.6596                                           \\ \hline
8                            & 0.9919                                                  & 0.5638                                           \\ \hline
\end{tabular}
\end{table}

Since the point of DrLIM is to preserve the similarity in a lower dimensional space, we further visualize the 2D mapping of 200 adversarial samples in Fig.~\ref{fig:DrLIM}(b). Note that these adversarial samples are generated from those 200 legitimate samples selected before. The 2D mapping in this case is not as clear as that for legitimate samples, especially since the overlapping is more obvious. However, the similarity between pairs of samples are still reasonably well-preserved. In order to explore more of these outliers, in Table~\ref{tab:DrLIMvsdnn}, we show the probabilities of making wrong classification decisions when testing a normal DNN and a DrLIM-DNN with these outliers. As shown in Table~\ref{tab:DrLIMvsdnn}, these outliers cause a normal DNN to make wrong classification results with over $97\%$ confidence. However, when processed with DrLIM-DNN, although these outliers are not mapped to ideal regions, the probabilities of being wrongly classified is significantly reduced to lower than $66\%$. This result indicates that a DrLIM-DNN is effective for responding to unfamiliar samples with lower confidence. Therefore, DrLIM-DNN will not suffer as much as a normal DNN would when confronted with highly confusing adversarial samples.

As our experimental results show, DrLIM-DNN provides the best performance when tested against adversarial samples. In particular, the DrLIM-DNN achieves a classification error rate $44.93\%$ (smaller than the classification error rate obtained using adversarial training). In addition, DrLIM-DNN also shows its effectiveness for avoiding making strongly confident predictions for adversarial samples. With respect to this property, there has been limited research that shows similar results. However, since the DrLIM requires also building a CNN model, the computational cost is much higher compared to both DLM-DNN and adversarial training. Therefore, to best utilize the advantages of DrLIM-DNN, it might be reasonable to use a smaller data set with a fewer number of classes.  

\subsection{Reconstruction Performance}
As previously introduced in Section \ref{sec:DRmethods}, both DLM-DNN and DrLIM-DNN are non-invertible for different reasons. More importantly, we have proven that recovering the original data from a low dimensional space induced by DrLIM is an NP-hard problem. In this subsection, we mainly focus on inverting the proposed dimensional reduction method DLM by approximating it with a linear transformation matrix. We obtain the linear transformation matrix by solving a linear regression problem. In case the original data is sparse, we further employ a linear regression with $L_1$ regularization. First, we demonstrate that when configuring DLM as pure PCA, the approach is not robust given that it may be effectively inverted and thus allow for reconstruction of adversarial samples. Next, we examine the reconstruction error obtained from inverting DLM, taking a percentage of PCA mappings less than $100\%$. 
\begin{figure*}[t]
    \centering
    \begin{tabular}{cc}
        \includegraphics[width=0.30\textwidth]{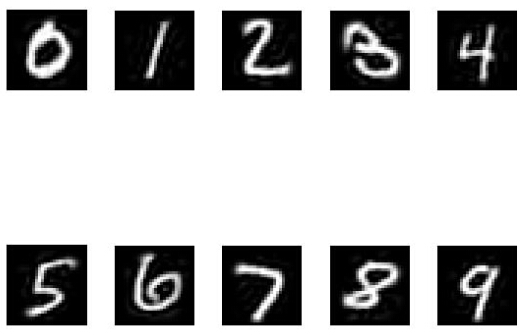} & \includegraphics[width=0.30\textwidth]{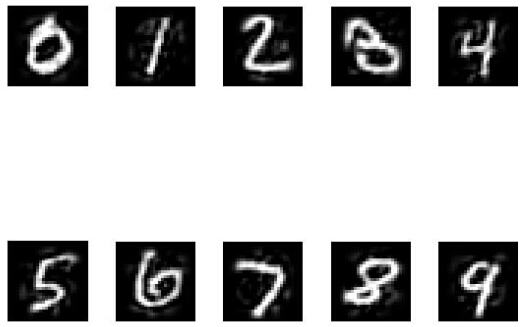} \\
        (a) Reconstructed legitimate samples & (b) Reconstructed  adversarial samples \\[6pt]
    \end{tabular}
    \caption{Reconstructed samples by inverting PCA-DNN. The reconstructed legitimate and adversarial samples are shown in (a) and (b) respectively. Specifically, the former are reconstructed based on the lower dimensional mappings of legitimate samples. As for the latter, reconstruction is conducted based on lower dimensional adversarial mappings.}
    \label{fig:inverted_PCA}
\end{figure*}

We evaluate the reconstruction performance when inverting one extreme case of DLM, where DLM uses only PCA mappings. In this way, the original samples are only processed by PCA before being input to a DNN model. We refer to this method as PCA-DNN for comparison. Therefore, an adversary can easily recover the adversarial images given their lower dimensional mappings. To examine this extreme case, we first configure DLM as pure PCA and map legitimate testing samples to a $100$-dimensional space. Then we reconstruct these legitimate samples by inverting PCA, as explained in Section \ref{sec:DRmethods}. To demonstrate the effectiveness of inverting PCA, we show reconstructed legitimate samples in Fig.~\ref{fig:inverted_PCA}(a). In this case, the reconstruction was successful  and the samples are haradly different from legitimate ones. 
\begin{table}[b]
\centering
\caption{Classification performance of PCA-DNN and DLM-DNN testing with reconstructed adversarial samples by inverting PCA }
\label{tab:pca_ad}
\begin{tabular}{@{}ccc@{}}
\toprule
\multicolumn{2}{c}{\multirow{2}{*}{\begin{tabular}[c]{@{}c@{}}Trained \\ models\end{tabular}}} & \multirow{2}{*}{\begin{tabular}[c]{@{}c@{}}Classification error rates of testing\\ with reconstructed adversarial samples\end{tabular}} \\
\multicolumn{2}{c}{}                                                                                                          &                                                                                                                                                          \\ \midrule
\multicolumn{2}{c}{\begin{tabular}[c]{@{}c@{}}Normal DNN \end{tabular}}                  & 0.6596                                                                                                                                                   \\ \midrule
\multirow{5}{*}{\begin{tabular}[c]{@{}c@{}}Noise\\ Coefficient of 0.1\end{tabular}}               & PCA(95\%)              & 0.2846                                                                                                                                                   \\ \cmidrule(l){2-3}
                                                                                                 & PCA(75\%)             & 0.2011                                                                                                                                                   \\ \cmidrule(l){2-3}
                                                                                                 & PCA(50\%)              & 0.1447                                                                                                                                                   \\ \cmidrule(l){2-3}
                                                                                                 & PCA(25\%)              & 0.1131                                                                                                                                                   \\ \cmidrule(l){2-3}
                                                                                                 & PCA(5\%)               & 0.3691                                                                                                                                                   \\ \midrule
\multirow{5}{*}{\begin{tabular}[c]{@{}c@{}}Noise\\ Coefficient of 0.3\end{tabular}}               & PCA(95\%)               & 0.1864                                                                                                                                                   \\ \cmidrule(l){2-3}
                                                                                                 & PCA(75\%)              & 0.1729                                                                                                                                                   \\ \cmidrule(l){2-3}
                                                                                                 & PCA(50\%)              & 0.1449                                                                                                                                                   \\ \cmidrule(l){2-3}
                                                                                                 & PCA(25\%)              & 0.1884                                                                                                                                                   \\ \cmidrule(l){2-3}
                                                                                                 & PCA(5\%)               & 0.4766                                                                                                                                                   \\ \bottomrule
\end{tabular}
\end{table}
Now we assume that an adversary has acquired the lower dimensional mappings generated by PCA. According to Section \ref{sec:background}, this adversarial can easily generate their corresponding \emph{lower dimensional adversarial mappings}. Following the inversion procedure of Section \ref{sec:DRmethods}, this adversary can further reconstruct adversarial samples based on these lower dimensional adversarial mappings. We also show the reconstructed adversarial samples in Fig.~\ref{fig:inverted_PCA}(b). While Fig.~\ref{fig:inverted_PCA}(a) and Fig.~\ref{fig:inverted_PCA}(b) are difficult to differentiate, we use the reconstructed adversarial samples to test a normal DNN model and a DLM-DNN under different settings. According to the testing results shown in Table~\ref{tab:pca_ad}, the reconstructed adversarial samples maintain their attack power against a normal DNN model. The classification error rate in this case is larger than $60\%$. In contrast, these adversarial samples can be effectively defended by a DLM-DNN as shown in Table~\ref{tab:DLM_DrLIM_DNN}.

\begin{figure}[t]
    \centering
    \includegraphics[width=0.38\textwidth]{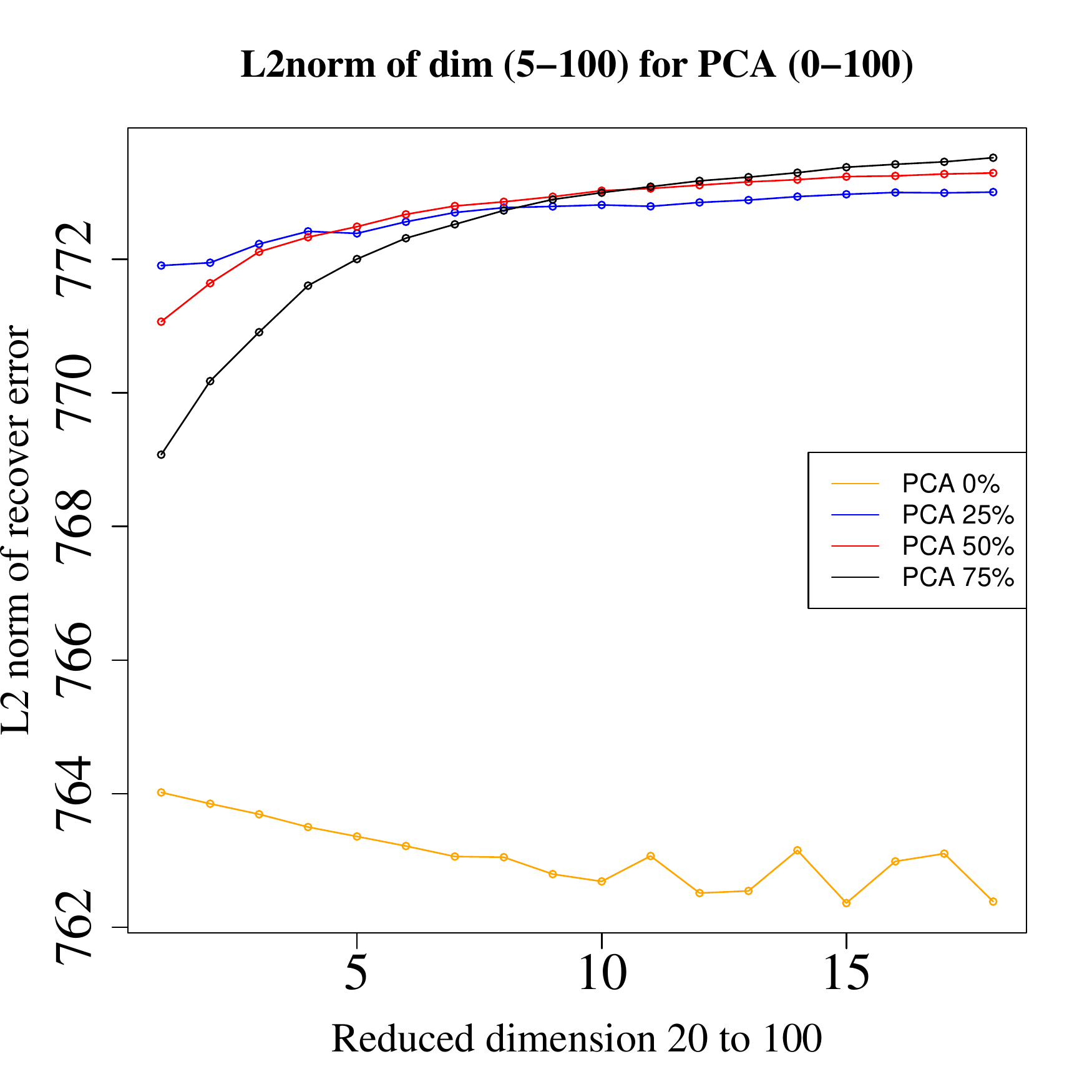} \\
    \caption{Reconstruction errors with varying parameters for inverting DLM. For each curve, it represents the reconstruction errors generated by setting reduced dimensions in the range of $(20, 100)$. Meanwhile, the percentage of PCA mapping is fixed to any value of $5\%$, $25\%$, $50\%$, $75\%$ and $95\%$.}
    \label{fig:recover_err}
\end{figure}

We finally investigate the reconstruction errors when inverting DLM-DNN. Note that since we have shown that using PCA is not secure for defending against reconstructed adversarial samples, the reconstruction errors for this extreme case will not included in following discussion. We present reconstruction errors when varying percentages of PCA mappings used and when varying sub-space dimensionality in Fig~\ref{fig:recover_err}. Our experiment shows that inverting a DLM-DNN leads to high reconstruction errors, regardless of how many PCA mappings are used what dimensionality is used. Recall the theoretical analysis of DrLIM-DLM in Section~\ref{sec:DRmethods}, we demonstrate that our proposed methods effectively build an adversary-resistant DNN.

\section{Related Work}
\label{sec:rw}
In this work, we build adversary-resilient DNN architectures using non-invertible dimension reduction methods. Related research commonly falls into the area of adversarial machine learning or dimensionality reduction methods. In this section we introduce several state-of-the-art adversarial machine learning technologies. These technologies can be categorized as either data augmentation or DNN model-complexity enhancement. Finally, we present several prevalent dimensionality reduction methods and study their suitability under our proposed framework.

\subsection{Adversarial Machine Learning}
Approaches to the robustness issue mainly focus on adversarial training and enhancing DNN complexity.\footnote{For details of robustness issues, please see Section \ref{sec:background}.}

Adversarial training approaches usually augment the training set by integrating adversarial samples~\cite{42503}. More formally, this data augmentation approach can be theoretically interpreted regularizing the training objective function. This regularization is designed to penalize certain gradients using the derivative of the loss with respect to the input variables. The penalized gradients indicate a direction in the input space that the objective function is most sensitive to. For example, recent work\cite{Goodfellow14} introduced such an objective function, directly combining a regularization term with the original cost function. Instead of training a DNN with a mixture of adversarial and legitimate samples, other approaches ~\cite{nokland2015improving} use only adversarial samples. Despite the difference between these strategies, a unifying framework,\emph{DataGrad}~\cite{ororbia_ii_unifying_2016}, was proposed to generalize adversarial training of deep architectures and help explain prior approaches~\cite{gu2014towards, rifai2011contractive}. 

However, since adversarial training still falls into the category of data augmentation, which, as explained earlier in this paper, cannot possibly hope to cover the entire infinite space of possible adversarial samples. In other words, adversarial training is still vulnerable to unforeseen adversarial samples. Therefore, we looked to improving the robustness of DNNs by enhancing the model complexity. 

Recent work~\cite{papernot2015distillation}, denoted as \emph{distillation} enhances a normal DNN by enabling it to utilize other informative inputs. More specifically, for each input training sample, instead of using its original \emph{hard label} which only indicates whether it belongs to a particular class, the enhanced DNN utilizes a \emph{soft label}. This means that probability distribution over all classes is used as a corresponding target vector. In order to use this more informative soft label, two DNNs must be designed successively and then stacked together. One DNN is designed for generating more informative input samples while the other DNN is trained using these more informative samples in the usual way. However, as discussed in section~\ref{sec:motivation}, once the architecture of distillation is disclosed to attackers, it is always possible to find adversarial samples based on this very defense mechanism. If an attacker~\cite{carlini2016defensive} generates adversarial samples by artificially reducing the magnitude of the input to the final softmax function and modifies the objective function used for generating adversarial samples~\cite{papernot2016limitations}, the defensively distilled networks are open to assault. A denoising autoencoder~\cite{gu2014towards} approach was also proposed to filter out adversarial perturbations. However, it too suffers from the same problem as~\cite{papernot2015distillation}. Again, once the architecture of the stacked DNN and auto-encoder is disclosed, attackers can still generate adversarial samples by treating this new model as a normal feed-forward DNN only with more hidden layers. Indeed, adversarial samples of this stacked DNN~\cite{gu2014towards} show even smaller distortion compared with adversarial samples generated from the original DNN. Thus, a successful method for increasing model complexity requires the property that even if the model architecture is disclosed, it would not be possible to generate adversarial samples based on this information.

% AO: not sure if we actually need this little paragraph...redundant?
%The above approaches do not yield completely adversarial-robust architectures. Therefore, we propose an adversary resilient DNN architecture that is not only robust against adversarial samples, but also robust against the mechanism of generating unseen adversarial samples specific to our proposed architecture.

\subsection{Dimensionality Reduction Methods}

From the previous analysis in Section~\ref{sec:framework}, we see that dimensionality reduction has definite advantages for the design goals of our proposed architecture using non-invertible data transformations. As such, we investigated various dimensional reduction methods including both linear and nonlinear methods. 

Recall from Section \ref{sec:DRmethods}, dimensionality reduction methods that preserve the similarity between high dimensional samples in the mapped lower dimensional space can be helpful in defending against adversarial samples. This is due to the fact that adversarial samples are highly similar to legitimate samples yet can be wrongly classified in a lower dimensional space. One popular linear dimension reduction method with aforementioned characteristic is multi-dimensional scaling (MDS)~\cite{kruskal1964nonmetric}. MDS is computationally efficient and has been widely adopted in various machine learning tasks. More importantly, MDS generates lower dimensional mappings based on similarity, which is usually represented by Euclidean distance, between high dimensional data points. Given this property, an adversarial sample will not be mapped that far away from its similar, legitimate neighbors. However, MDS have three shortcomings which restrict its usage in our proposed framework. First, since MDS is built only with distance information between samples, critical feature information contained in data samples cannot be preserved. Second, MDS cannot provide a mapping from high dimensional input to a lower dimensional output when being applied to unseen data samples. When stacking DNN on top of MDS, whenever unseen samples appears, the entire combined architecture needs to be retrained, which is computationally undesirable.

% AO: wait what? why are you saying your proposed methods suffer the weaknesses of MDS? This would destroy the entire paper... I commented out he below phrase from the end of the last paragraph
%and the DLM and DrLIM introduced in Section \ref{sec:DRmethods}.

We also investigated another nonlinear dimensional reduction methods of t-SNE~\cite{maaten2008visualizing}. Specially, t-SNE is similar to MDS in that it is also capable of preserving the similarity between high dimensional data in a lower dimensional space. t-SNE minimizes KL-divergence between two probability distributions, one representing the probability of being neighbors in a high dimensional space while the other represents the probability of being neighbors in a lower dimensional space. However, similar to MDS, t-SNE is not suitable for online processing. 
%Otherwise unseen samples will lead to the consequence of recomputing both t-SNE and DNN models.

The dimensionality reduction methods used in our proposed framework take these aforementioned issues into account. In particular, DLM is more computationally efficient in that it can easily preserve feature information of the original data samples. Meanwhile, DrLIM not only preserves similarity and important feature information of original input space simultaneously, but is an incremental algorithm. More importantly, both DLM and DrLIM are non-invertible which we justify in theoretically and empirically in Sections \ref{sec:DRmethods} and \ref{sec:eval}. 
%Therefore, by integrating either dimension method into our framework, we can build an adversary resilient DNN architecture.
\section{Conclusion}
\label{sec:conclusion}

We proposed a new framework for constructing deep neural network models that are robust to adversarial samples. Our framework design is based on an analysis of both the ``blind-spot'' of DNNs and the limitations of currently proposed solutions. 

Using our proposed framework, we developed two adversary-resilient DNN architectures that leverage non-invertible data transformation mechanisms. Using the first proposed approach for processing the data fed into DNN models, we empirically showed that crafting an adversarial sample for this architecture will incur significant distortion and thus lead to easily detectable adversarial samples. In contrast, under the second architecture, we theoretically demonstrated that it is impossible for an adversary to craft an adversarial sample to attack it. This implies that our proposed framework no longer suffers from attacks that rely on generating model-specific adversarial samples. 

Furthermore, we demonstrated that recently studied adversarial training methods are not sufficient defense mechanisms. Applying our new framework to the MNIST data set, we empirically demonstrate that our new framework significantly reduces the error rates in classifying adversarial samples. Furthermore, our new framework has the same classification performance for legitimate samples with negligible degradation. Future work will entail investigating the performance of our framework in a wider variety of applications.

% conference papers do not normally have an appendix

% use section* for acknowledgement
%\section*{Acknowledgment}
%The authors would like to thank...

\ifCLASSOPTIONcaptionsoff
  \newpage
\fi

\bibliography{ref}

\begin{thebibliography}{10}

\bibitem{bar2015deep}
Y.~Bar, I.~Diamant, L.~Wolf, and H.~Greenspan.
\newblock Deep learning with non-medical training used for chest pathology
  identification.
\newblock In {\em SPIE Medical Imaging}, pages 94140V--94140V. International
  Society for Optics and Photonics, 2015.

\bibitem{Goodfellow-et-al-2016-Book}
I.~G.~Y. Bengio and A.~Courville.
\newblock Deep learning.
\newblock Book in preparation for MIT Press, 2016.

\bibitem{carlini2016defensive}
N.~Carlini and D.~Wagner.
\newblock Defensive distillation is not robust to adversarial examples.
\newblock {\em arXiv preprint arXiv:1607.04311}, 2016.

\bibitem{dahl2013large}
G.~E. Dahl, J.~W. Stokes, L.~Deng, and D.~Yu.
\newblock Large-scale malware classification using random projections and
  neural networks.
\newblock In {\em 2013 IEEE International Conference on Acoustics, Speech and
  Signal Processing}, pages 3422--3426. IEEE, 2013.

\bibitem{d2003relaxations}
A.~d?Aspremont and S.~Boyd.
\newblock Relaxations and randomized methods for nonconvex qcqps.
\newblock {\em EE392o Class Notes, Stanford University}, 2003.

\bibitem{Compressed2006Donoho}
D.~L. Donoho.
\newblock Compressed sensing.
\newblock {\em IEEE Transactions on Information Theory}, 52(4):1289--1306,
  April 2006.

\bibitem{eckart1936approximation}
C.~Eckart and G.~Young.
\newblock The approximation of one matrix by another of lower rank.
\newblock {\em Psychometrika}, 1(3):211--218, 1936.

\bibitem{farabet2012scene}
C.~Farabet, C.~Couprie, L.~Najman, and Y.~LeCun.
\newblock Scene parsing with multiscale feature learning, purity trees, and
  optimal covers.
\newblock {\em arXiv preprint arXiv:1202.2160}, 2012.

\bibitem{fornasier2011compressive}
M.~Fornasier and H.~Rauhut.
\newblock Compressive sensing.
\newblock In {\em Handbook of mathematical methods in imaging}, pages 187--228.
  Springer, 2011.

\bibitem{Goodfellow14}
I.~Goodfellow, J.~Shlens, and C.~Szegedy.
\newblock {Explaining and Harnessing Adversarial Examples}.
\newblock {\em CoRR}, 2014.

\bibitem{gu2014towards}
S.~Gu and L.~Rigazio.
\newblock Towards deep neural network architectures robust to adversarial
  examples.
\newblock {\em {arXiv}:1412.5068 [cs]}, 2014.

\bibitem{Hadsell2006Dimensionality}
R.~Hadsell, S.~Chopra, and Y.~LeCun.
\newblock Dimensionality reduction by learning an invariant mapping.
\newblock In {\em 2006 IEEE Computer Society Conference on Computer Vision and
  Pattern Recognition (CVPR'06)}, volume~2, pages 1735--1742, 2006.

\bibitem{hadsell2009learning}
R.~Hadsell, P.~Sermanet, J.~Ben, A.~Erkan, M.~Scoffier, K.~Kavukcuoglu,
  U.~Muller, and Y.~LeCun.
\newblock Learning long-range vision for autonomous off-road driving.
\newblock {\em Journal of Field Robotics}, 26(2):120--144, 2009.

\bibitem{hinton2007learning}
G.~E. Hinton.
\newblock Learning multiple layers of representation.
\newblock {\em Trends in cognitive sciences}, 11(10):428--434, 2007.

\bibitem{huang2015learning}
R.~Huang, B.~Xu, D.~Schuurmans, and C.~Szepesv{\'a}ri.
\newblock Learning with a strong adversary.
\newblock {\em CoRR, abs/1511.03034}, 2015.

\bibitem{jolliffe2002principal}
I.~Jolliffe.
\newblock {\em Principal component analysis}.
\newblock Wiley Online Library, 2002.

\bibitem{kruskal1964nonmetric}
J.~B. Kruskal.
\newblock Nonmetric multidimensional scaling: a numerical method.
\newblock {\em Psychometrika}, 29(2):115--129, 1964.

\bibitem{lecun1998mnist}
Y.~LeCun, C.~Cortes, and C.~J. Burges.
\newblock The mnist database of handwritten digits, 1998.

\bibitem{maaten2008visualizing}
L.~v.~d. Maaten and G.~Hinton.
\newblock Visualizing data using t-sne.
\newblock {\em Journal of Machine Learning Research}, 9(Nov):2579--2605, 2008.

\bibitem{miyato2015distributional}
T.~Miyato, S.-i. Maeda, M.~Koyama, K.~Nakae, and S.~Ishii.
\newblock Distributional smoothing with virtual adversarial training.
\newblock {\em stat}, 1050:25, 2015.

\bibitem{ngiam2011optimization}
J.~Ngiam, A.~Coates, A.~Lahiri, B.~Prochnow, Q.~V. Le, and A.~Y. Ng.
\newblock On optimization methods for deep learning.
\newblock In {\em Proceedings of the 28th International Conference on Machine
  Learning (ICML-11)}, pages 265--272, 2011.

\bibitem{nguyen2015deep}
A.~Nguyen, J.~Yosinski, and J.~Clune.
\newblock Deep neural networks are easily fooled: High confidence predictions
  for unrecognizable images.
\newblock In {\em 2015 IEEE Conference on Computer Vision and Pattern
  Recognition (CVPR)}, pages 427--436. IEEE, 2015.

\bibitem{nokland2015improving}
A.~N{\o}kland.
\newblock Improving back-propagation by adding an adversarial gradient.
\newblock {\em {arXiv}:1510.04189 [cs]}, 2015.

\bibitem{ororbia_ii_unifying_2016}
A.~G. Ororbia~{II}, C.~L. Giles, and D.~Kifer.
\newblock Unifying adversarial training algorithms with flexible deep data
  gradient regularization.
\newblock {\em {arXiv}:1601.07213 [cs]}, 2016.

\bibitem{papernot2016limitations}
N.~Papernot, P.~McDaniel, S.~Jha, M.~Fredrikson, Z.~B. Celik, and A.~Swami.
\newblock The limitations of deep learning in adversarial settings.
\newblock In {\em 2016 IEEE European Symposium on Security and Privacy
  (EuroS\&P)}, pages 372--387. IEEE, 2016.

\bibitem{papernot2015distillation}
N.~Papernot, P.~McDaniel, X.~Wu, S.~Jha, and A.~Swami.
\newblock Distillation as a defense to adversarial perturbations against deep
  neural networks.
\newblock {\em arXiv preprint arXiv:1511.04508}, 2015.

\bibitem{pascanu2015malware}
R.~Pascanu, J.~W. Stokes, H.~Sanossian, M.~Marinescu, and A.~Thomas.
\newblock Malware classification with recurrent networks.
\newblock In {\em 2015 IEEE International Conference on Acoustics, Speech and
  Signal Processing (ICASSP)}, pages 1916--1920. IEEE, 2015.

\bibitem{raskutti2011minimax}
G.~Raskutti, M.~J. Wainwright, and B.~Yu.
\newblock Minimax rates of estimation for high-dimensional linear regression
  over-balls.
\newblock {\em IEEE Transactions on Information Theory}, 57(10):6976--6994,
  2011.

\bibitem{rifai2011contractive}
S.~Rifai, P.~Vincent, X.~Muller, X.~Glorot, and Y.~Bengio.
\newblock Contractive auto-encoders: Explicit invariance during feature
  extraction.
\newblock In {\em Proceedings of the 28th international conference on machine
  learning (ICML-11)}, pages 833--840, 2011.

\bibitem{roweis2000nonlinear}
S.~T. Roweis and L.~K. Saul.
\newblock Nonlinear dimensionality reduction by locally linear embedding.
\newblock {\em Science}, 290(5500):2323--2326, 2000.

\bibitem{rumelhart1988learning}
D.~E. Rumelhart, G.~E. Hinton, and R.~J. Williams.
\newblock Learning representations by back-propagating errors.
\newblock {\em Cognitive modeling}, 5(3):1, 1988.

\bibitem{shin2015recognizing}
E.~C.~R. Shin, D.~Song, and R.~Moazzezi.
\newblock Recognizing functions in binaries with neural networks.
\newblock In {\em 24th USENIX Security Symposium (USENIX Security 15)}, pages
  611--626, 2015.

\bibitem{silver2016mastering}
D.~Silver, A.~Huang, C.~J. Maddison, A.~Guez, L.~Sifre, G.~Van Den~Driessche,
  J.~Schrittwieser, I.~Antonoglou, V.~Panneershelvam, M.~Lanctot, et~al.
\newblock Mastering the game of go with deep neural networks and tree search.
\newblock {\em Nature}, 529(7587):484--489, 2016.

\bibitem{srivastava2014dropout}
N.~Srivastava, G.~E. Hinton, A.~Krizhevsky, I.~Sutskever, and R.~Salakhutdinov.
\newblock Dropout: a simple way to prevent neural networks from overfitting.
\newblock {\em Journal of Machine Learning Research}, 15(1):1929--1958, 2014.

\bibitem{42503}
C.~Szegedy, W.~Zaremba, I.~Sutskever, J.~Bruna, D.~Erhan, I.~Goodfellow, and
  R.~Fergus.
\newblock Intriguing properties of neural networks.
\newblock In {\em International Conference on Learning Representations}, 2014.

\bibitem{Vavasis1991Nonlinear}
S.~A. Vavasis.
\newblock {\em Nonlinear Optimization: Complexity Issues}.
\newblock Oxford University Press, Inc., New York, NY, USA, 1991.

\bibitem{xu2014deep}
Y.~Xu, T.~Mo, Q.~Feng, P.~Zhong, M.~Lai, I.~Eric, and C.~Chang.
\newblock Deep learning of feature representation with multiple instance
  learning for medical image analysis.
\newblock In {\em 2014 IEEE International Conference on Acoustics, Speech and
  Signal Processing (ICASSP)}, pages 1626--1630. IEEE, 2014.

\bibitem{yuan2014droid}
Z.~Yuan, Y.~Lu, Z.~Wang, and Y.~Xue.
\newblock Droid-sec: Deep learning in android malware detection.
\newblock In {\em ACM SIGCOMM Computer Communication Review}, volume~44, pages
  371--372. ACM, 2014.

\bibitem{zeiler2014visualizing}
M.~D. Zeiler and R.~Fergus.
\newblock Visualizing and understanding convolutional networks.
\newblock In {\em European Conference on Computer Vision}, pages 818--833.
  Springer, 2014.

\end{thebibliography}
\bibliographystyle{abbrv}

% that's all folks
\end{document}